\pdfoutput=1
\documentclass[12pt]{article}

\usepackage[margin=1in]{geometry}
\usepackage{amsmath,amssymb}
\usepackage{mathptmx}            
\usepackage{setspace}
\usepackage{booktabs}
\usepackage{xcolor}
\usepackage{colortbl}
\usepackage{array}
\usepackage[authoryear,round]{natbib}
\usepackage[hidelinks]{hyperref}
\usepackage{enumitem}
\usepackage{graphicx}
\usepackage{tikz}
\usetikzlibrary{arrows.meta,positioning,fit,backgrounds}

\usepackage{lineno}
\newcommand*\patchAmsMathEnvironmentForLineno[1]{%
  \expandafter\let\csname old#1\expandafter\endcsname\csname #1\endcsname
  \expandafter\let\csname oldend#1\expandafter\endcsname\csname end#1\endcsname
  \renewenvironment{#1}%
    {\linenomath\csname old#1\endcsname}%
    {\csname oldend#1\endcsname\endlinenomath}}
\newcommand*\patchBothAmsMathEnvironmentsForLineno[1]{%
  \patchAmsMathEnvironmentForLineno{#1}%
  \patchAmsMathEnvironmentForLineno{#1*}}
\AtBeginDocument{%
  \patchBothAmsMathEnvironmentsForLineno{equation}%
  \patchBothAmsMathEnvironmentsForLineno{align}%
  \patchBothAmsMathEnvironmentsForLineno{gather}%
  \patchBothAmsMathEnvironmentsForLineno{multline}}

\newcommand{\SE}{SE(3)}
\newcommand{\se}{\mathfrak{se}(3)}
\newcommand{\Ad}{\mathrm{Ad}}
\newcommand{\ad}{\mathrm{ad}}
\newcommand{\obj}{o}
\newcommand{\R}{\mathbb{R}}
\newcounter{algo}
\renewcommand{\thealgo}{\arabic{algo}}

\title{\textbf{Reference frames are the static limit of a dynamical representation}\\[2pt]
\large Intuitive physics on a sensorimotor reference-frame substrate}
\author{Stefano Goria\\[2pt]\small AIM Research Lab, \url{https://aimresearchlab.com}\\
\small Kinous Limited, London, UK\\[2pt]
\small \texttt{stefano@aimresearchlab.com}}
\date{November 2026}

\begin{document}
\singlespacing
\maketitle
\begin{abstract}
Reference-frame theories of object recognition and simulation theories of intuitive physics have
developed separately. In the first, an object is recognized by matching what is sensed to locations in
a coordinate frame attached to the object; in the second, its motion is predicted by running an
internal model of mechanics forward. For rigid objects we read the two as one representation. A
recognized reference frame gives an object's position and orientation; adding to the same frame its
velocity, the forces on it and its mass distribution gives the laws of rigid-body motion, and a single
transformation set by the object's pose carries all of these between frames. We call this the
dynamical completion of the reference frame. The law of motion is the same for every object; what
differs between objects, how their mass is distributed, is inferred by accumulating evidence, as pose
is. We built the completion into Monty, the sensorimotor recognition system of the Thousand Brains
Project: when an object moves, how well each candidate identity and pose explains the motion counts as evidence for that candidate, while recognition of objects at rest is unchanged. We test it on objects with
identical surfaces and different hidden masses, which touch and vision cannot tell apart. Thrown a few
times and observed in flight, such an object is identified. We compare the completion with the same
law of motion run as a separate module that reads the recognizer's candidates once; the completion
identifies objects more reliably. In a second experiment the system is not told that such look-alike
objects differ, and discovers from their motion that they do. The framework also makes a prediction
for neural recordings. A network of neuron-like units can compute the transformation of velocity
between frames, and it predicts that velocity tuning in an object's frame rotates with the object's
recognized pose.
\vspace{6pt}
\end{abstract}

\section{Introduction}
\label{sec:intro}
Two research programmes describe how brains represent objects and how they predict what objects will
do. Reference-frame theories propose that
cortex recognizes an object by matching sensed features to locations in a coordinate frame attached
to the object \citep{hawkins2017, tbt2019, lewis2019}. Monty, the implementation of the Thousand Brains
Theory, recognizes objects and their pose from a few touches or glances in this way: each of its
learning modules, modelled on a cortical column, learns complete models of objects in their own
reference frames and accumulates evidence for hypotheses about the object it is sensing, and several
modules reach a consensus by voting \citep{monty2024, tbs2026}. Simulation theories of intuitive physics propose that people predict
physical outcomes by running an approximate Newtonian model forward, with object properties inferred
from observation \citep{battaglia2013, hamrick2016, ullman2018, smith2024}. Reference-frame theories have so far addressed objects at rest. In simulation theories, objects and their pose usually come from a separate perceptual stage,
although physics can enter perception: in analysis-by-synthesis models of shape perception, the
inferred shape of a hidden object must explain how it deforms a cloth draped over it
\citep{yildirim2024}.

\begin{figure}[t]\centering
\includegraphics[width=0.85\linewidth]{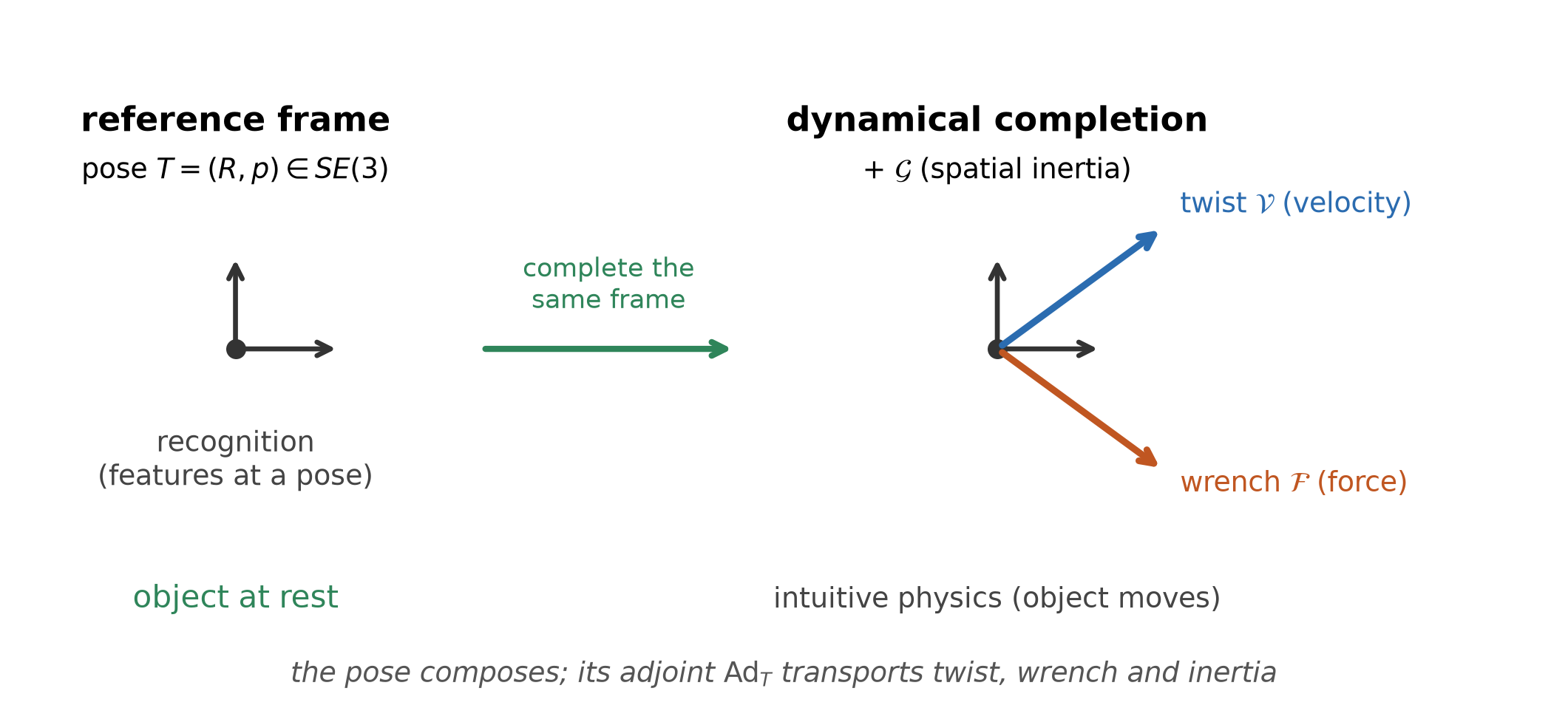}
\caption{The dynamical completion. A reference-frame recognizer represents the pose $T\in\SE$ of an
object at rest. Completing the same frame with a twist $\mathcal{V}$, a wrench $\mathcal{F}$ and a
spatial inertia $\mathcal{G}$ yields rigid-body dynamics. Poses compose by the group operation; the
adjoint $\Ad_T$ of the pose transports twists, wrenches and inertia between frames.}\label{fig:concept}
\end{figure}

We propose to read the two, for rigid bodies, as parts of one structure. A recognized reference frame gives the pose of an
object, its position and orientation, which is an element of the group of rigid motions. The object's
velocity, the force acting on it and its mass distribution can be attached to the same frame, and one
linear map determined by the pose moves all of them between frames
\citep{featherstone2008, lynchpark2017}. We call this extension of the recognized frame the
\emph{dynamical completion} of the reference frame (Fig.~\ref{fig:concept}). It yields rigid-body
dynamics with a law that is the same for every object and per-object parameters that can be inferred
the way pose is inferred.

The completion connects to recognition through the hypotheses that each of Monty's learning modules holds about which object it is sensing and how that object is oriented, and for which the module accumulates evidence as it senses. The completion extends this part of Monty, the hypotheses of a learning module, and leaves voting between modules as it is. In the completion the dynamics share
these hypotheses: the likelihood of an
observed motion under the law becomes evidence for or against each of them. This matters when objects look alike and move differently, such as a full and an empty bottle, or a box with something heavy inside: their surfaces are the same, so touch and vision cannot tell them apart, but they move differently.

The mathematics is the standard spatial-vector form of rigid-body mechanics, and our contribution is to organize it inside a reference-frame recognizer. The group fixes how velocity, force and inertia transform between frames, while the parametrization of inertia, the inference algorithm and any cortical implementation are our choices, which we state as such.

We implemented the completion in Monty and ask three questions of it. First, does it leave Monty's
recognition unchanged, and does its law behave as a physical law should? Four checks (C1--C4) confirm
that recognition is unchanged and that the law learns an object's dynamics from a few throws, transfers
across pose and adapts to hidden masses. Second, among objects that look the same and move differently,
does motion that enters the recognizer's hypotheses improve recognition? Experiment~1 (E1) tests this
and compares the completion with the same law run as a separate module attached to the recognizer from
outside; the comparison shows which benefits depend on access to the hypotheses as sensing revises
them. Third, does motion reveal, without labels, that one surface covers several objects? Experiment~2
(E2) tests this. Finally, we ask how neurons could carry out the completion: a network of neuron-like
units computes its central operation, the transport of velocity between frames, and yields predictions
for neural recordings.

\section{Background: Monty and the Thousand Brains Theory}
\label{sec:tbm}
\label{sec:static}
This section summarizes the parts of Monty and of the Thousand Brains Theory that the completion builds
on, first as concepts and then in the notation used in the rest of the paper.

\subsection{Concepts: columns, reference frames, prediction and voting}
\label{sec:tbmconcepts}
The neocortex consists of many similar columns, each receiving input from a small region of a sensory
surface, and the Thousand Brains Theory proposes that each column learns complete models of objects
\citep{hawkins2017}. Monty implements a column as a \emph{learning module}, which receives input from
one sensor patch \citep{monty2024, tbs2026}. The sensor moves over the object, and after each movement,
which we call a \emph{step}, the patch reports its location and the features sensed there: the surface
normal and principal-curvature directions, which give the patch an orientation, the curvature
magnitudes and, where available, colour. A module stores an object as a set of features at locations in
a reference frame attached to the object. The theory also assigns these roles to the layers of the
cortical sheet: it proposes that grid-cell-like cells in layer 6a represent the location of the sensor
in the object's frame, and that layer 4 represents the sensed feature in the context of that location
\citep{lewis2019, tbt2019}.

To recognize an object, a module holds a set of hypotheses, each a candidate object at a candidate pose. When the sensor moves, each hypothesis implies a new location on its object and therefore what should be sensed there. A hypothesis whose prediction matches the input
gains evidence and one whose prediction fails loses it, so the set of candidates narrows with each step
until one object and pose remain \citep{hawkins2017, lewis2019}.

In the theory, a predicted input activates only the cells that predicted it, a sparse response, whereas an input that no cell predicted activates all the cells of the selected minicolumns; this denser response indicates an unanticipated input and activates every stored context in which that input can occur \citep{hawkins2016, lewis2019}. In Monty the corresponding mechanism measures how fast the evidence of the best hypothesis grows, and when that growth stalls, which its implementation calls a
high prediction error, it samples new hypotheses for a few steps \citep{tbpmonty}. The completion uses
this mechanism for motion: a motion that contradicts every hypothesis stalls their evidence and
triggers the same sampling.

In the theory, long-range connections between the output layers of columns that sense different parts of the same object let them reach a consensus on the object
after one or a few sensations, where a single column would need several movements
\citep{hawkins2017, tbt2019}. Monty implements this as \emph{voting}: each module sends its hypotheses
and their evidence to the others, the receiving module shifts each vote by the displacement between the
two sensors, and it raises the evidence of its own hypotheses that agree with the votes; with more
modules Monty recognizes objects in fewer steps \citep{monty2024, tbs2026}. The completion acts inside
a module and leaves voting unchanged.

\subsection{Formal summary}
\label{sec:tbmformal}
An object $\obj$ is stored as $N_\obj$ nodes $\{(\ell_k,n_k,U_k,\varphi_k)\}$: a location
$\ell_k\in\R^3$, a unit surface normal $n_k$, a tangent frame $U_k\in SO(3)$ (the normal completed by
the two principal-curvature directions) and non-pose features $\varphi_k$ (curvature magnitudes and
colour); $SO(3)$ is the group of rotations of three-dimensional space. A module maintains hypotheses
$h_i=(\obj_i,R_i,\ell_i)$, each a candidate object, a candidate orientation $R_i\in SO(3)$ and a
candidate location $\ell_i$ of the sensor on that object, with a scalar evidence $e_i$. When the
sensor moves by $d$ in the agent's body frame, each hypothesis moves its location and updates its
evidence,
\begin{equation}
\ell_i \leftarrow \ell_i + R_i^{\top} d,
\qquad
e_i \leftarrow w_{\text{p}}\,e_i + w_{\text{c}}\big(\text{ev}^{\text{feat}}_i + \text{ev}^{\text{pose}}_i\big),
\label{eq:evidence}
\end{equation}
where $\text{ev}^{\text{feat}}_i$ scores the feature match at the nearest stored node,
$\text{ev}^{\text{pose}}_i$ the alignment of the sensed and stored orientations under $R_i$, and
$w_{\text{p}},w_{\text{c}}$ are mixing weights. The evidence is bounded and unnormalized.

To detect prediction error, Monty tracks for each hypothesis the slope $s_i$, the mean change of $e_i$ per step over a sliding window of recent steps, and applies two thresholds \citep{tbpmonty}:
\begin{equation}
s_i<\theta_{\text{del}}\ \Rightarrow\ \text{delete } h_i,
\qquad
\max_i s_i\le\theta_{\text{burst}}\ \Rightarrow\ \text{sample new hypotheses for } B \text{ steps}.
\label{eq:burst}
\end{equation}
A hypothesis that stops gaining evidence is deleted, and when even the best one stops gaining, Monty adds hypotheses to every object for $B$ steps, which it calls a sampling burst; this is the analogue of the theory's dense response to an unanticipated input \citep{hawkins2016}.

In voting, a module $a$ sends each of its hypotheses $h_j$ whose evidence passes a threshold, with the
evidence rescaled to $\tilde e_j\in[-1,1]$ over its current hypotheses. If the sensor of module $b$ is
displaced by $d_{ab}$ from that of $a$, the vote is moved to where $b$'s sensor would be under the same
hypothesis, $\ell_j'=\ell_j+R_j^{\top}d_{ab}$. Module $b$ then adds to each of its hypotheses of the
same object the strongest vote among the nearest transported locations within a search radius,
\begin{equation}
e_i \leftarrow e_i + w_{\text{v}}\max_{j\in\mathcal N(\ell_i)}\tilde e_j,
\label{eq:vote}
\end{equation}
with $\mathcal N(\ell_i)$ those votes and $w_{\text{v}}$ the vote weight; when the evidence is a running average the sum is normalized. In the version of Monty used here \citep{tbpmonty}, the receiving module takes the strongest of the three nearest votes (the published description uses a distance-weighted average; \citealp{monty2024}) and compares locations only, although votes also carry the hypothesized orientation.

Monty treats as \emph{possible} the objects whose best evidence is within a set fraction of the overall best, and within each such object the poses within the same
fraction of its best. The output is the hypothesis with the highest evidence: an object and its pose
$T=(R,p)$, with position $p\in\R^3$, an element of $\SE$, the group of rigid motions of
three-dimensional space. A hypothesis-testing policy moves the sensor to the stored point that best
separates the two leading hypotheses \citep{tbpmonty}. Everything in this section concerns objects at
rest; the completion adds a term for motion to the evidence update \eqref{eq:evidence}, and that term
acts on recognition through the thresholds of \eqref{eq:burst}.

\section{The dynamical completion of a reference frame}
\label{sec:model}

\subsection{Dynamics on the recognized frame}
\label{sec:dyn}
Recognition estimates the pose $T\in\SE$ of an object in a reference frame, and the completion adds to that frame the object's velocity (the twist $\mathcal{V}$), the torque and force acting on it (the wrench $\mathcal{F}$) and its mass distribution (the spatial inertia $\mathcal{G}$),
\begin{equation}
\mathcal{V}=(\omega,v)\in\se,\quad
\mathcal{F}=(\tau,f)\in\se^{*},\quad
\mathcal{G}=\begin{bmatrix}\mathbf{I}_\obj & m\,[r]_\times\\ -m\,[r]_\times & m\mathbf{1}\end{bmatrix},
\label{eq:inertia}
\end{equation}
where $\se$ is the Lie algebra of $\SE$, the six-dimensional space of rigid-body velocities, and
$\se^{*}$ its dual, the space of torques and forces; $\omega,v\in\R^3$ are angular and linear velocity
in the body frame, $\tau,f$ torque and force, $m$ the mass, $r\in\R^3$ the centre-of-mass offset from
the frame origin, $[\,\cdot\,]_\times$ the cross-product matrix, $\mathbf{1}$ the $3\times3$ identity and
$\mathbf{I}_\obj$ the rotational inertia about the frame origin. The adjoint of the pose is the linear map that moves these three quantities between frames. For a pose $T=(R,p)$ it is
\begin{equation}
\Ad_T=\begin{bmatrix}R&0\\ [p]_\times R & R\end{bmatrix},\qquad
\mathcal{V}_a=\Ad_{T_{ab}}\mathcal{V}_b,\quad
\mathcal{F}_a=\Ad_{T_{ba}}^{\top}\mathcal{F}_b,\quad
\mathcal{G}_a=\Ad_{T_{ba}}^{\top}\mathcal{G}_b\,\Ad_{T_{ba}},
\label{eq:adjoint}
\end{equation}
where $T_{ab}$ is the pose of frame $b$ in frame $a$ and a subscript names the frame in which a
quantity is expressed. The pose that recognition infers is therefore the one that carries the object's
inertia into the world. With the matrix of the Lie bracket
$\ad_{\mathcal{V}}=\left[\begin{smallmatrix}[\omega]_\times&0\\ [v]_\times&[\omega]_\times\end{smallmatrix}\right]$,
the law of motion is
\begin{equation}
\mathcal{G}\dot{\mathcal{V}}-\ad_{\mathcal{V}}^{\top}(\mathcal{G}\mathcal{V})=\mathcal{F},
\qquad
\mathcal{F}=\mathcal{G}\,\Ad_T^{-1}\mathbf{a}_g + \mathcal{F}_{\text{medium}} + \mathcal{F}_{\text{contact}},
\qquad
T\leftarrow T\exp(\Delta t\,[\mathcal{V}]),
\label{eq:law}
\end{equation}
the Newton--Euler equations in the body frame, with gravity (the free-fall acceleration $\mathbf{a}_g$
brought into the body frame), drag $\mathcal{F}_{\text{medium}}$ when the object moves through a medium
and contact $\mathcal{F}_{\text{contact}}$ when it bounces (Appendix~\ref{sec:strataapp}),
and the update of the pose over a time step $\Delta t$, where $\exp$ is the matrix exponential and
$[\mathcal{V}]=\left[\begin{smallmatrix}[\omega]_\times & v\\ 0 & 0\end{smallmatrix}\right]$ the $4\times4$
matrix form of the twist. We integrate \eqref{eq:law} with a fourth-order Runge--Kutta--Munthe-Kaas scheme,
which keeps $T$ on $\SE$.

\subsection{Parameter inference}
\label{sec:fixedlearned}
The law \eqref{eq:law}, the $\SE$ representation and the adjoint are fixed and the same for every
object. What differs between objects, the parameters $\theta_\obj$, is inferred: mass, centre of mass and principal inertias (and drag coefficient and
restitution, when a medium or contact excites them), held as an unconstrained $9$-vector $z$
(Appendix~\ref{sec:methods}). In free flight
the centre of mass and the ratios of the principal inertias are determined, while absolute mass is not
(Appendix~\ref{sec:identifiability}). The parameters are inferred by evidence accumulation of the same
kind as pose, implemented as a particle filter with $J$ particles. An \emph{interaction} is one
observed throw: a launch pose $T_0$ and twist $\mathcal{V}_0$ and the recorded poses $\hat T_\tau$,
$\tau=1,\dots,N_\tau$. Particle $j$, with parameters $z_j$, is rolled out by the law from the launch state, giving poses
$T(\tau;z_j)$, and its log weight $\log w_j$ is decremented by its mean squared trajectory error,
\begin{equation}
\log w_j \leftarrow \log w_j - \frac{\bar d_j^{\,2}}{2\sigma^2},\qquad
\bar d_j^{\,2} = \frac{1}{N_\tau}\sum_{\tau}\Big(w_{\text{rot}}\big\lVert\log(T(\tau;z_j)^{-1}\hat T_\tau)_\omega\big\rVert^2
+ \big\lVert\log(T(\tau;z_j)^{-1}\hat T_\tau)_v\big\rVert^2\Big),
\label{eq:paramfilter}
\end{equation}
where $\log$ is the $\SE$ logarithm with rotational and translational parts $\omega,v$, $\sigma$ an
error tolerance and $w_{\text{rot}}$ the relative weight of attitude error; particles are resampled and
perturbed when the effective sample size falls below $J/2$ (Appendix~\ref{sec:methods}).

\subsection{The completion inside Monty}
\label{sec:attach}
We implement the completion as a subclass of
\texttt{EvidenceGraphLM}, Monty's evidence learning
module, that adds one operation, the \emph{motion step}, called when the sensed object moves. Monty's
hypotheses updater (its shipped burst sampler), evidence update and thresholds are used unchanged.
Each object's parameters $\theta_\obj$ are stored with its model.

A motion step first adds dynamical evidence to Monty's hypothesis store. A tracker of the surface provides the rigid displacement $\Delta(\tau)=\hat T(\tau)\hat T(0)^{-1}$ of the
moving body, its angular velocity $\omega_w$ at launch, and the position $c$ and velocity $v_c$ of one
tracked surface point, all in the world frame. A hypothesis $h=(\obj,R_h,\ell_h)$ asserts the object's
frame at launch: orientation $R_h$, and origin $p_h$ given by aligning all surface points sensed so far
with the stored model under $R_h$. Under $h$ the observed trajectory of the frame is
$\Delta(\tau)\,T_h(0)$ with $T_h(0)=(R_h,p_h)$, and the launch twist is
$\mathcal V_h(0)=\big(R_h^{\top}\omega_w,\ R_h^{\top}(v_c+\omega_w\times(p_h-c))\big)$. The law is
integrated from there with $\obj$'s parameters, and the hypothesis receives
\begin{equation}
\text{ev}^{\text{dyn}}_h=\max\!\Big(-\frac{\bar d_h^{\,2}}{2\sigma_{\text{dyn}}^2},\ -b_{\max}\Big),
\label{eq:dynevidence}
\end{equation}
with $\bar d_h$ the trajectory error of \eqref{eq:paramfilter}, $\sigma_{\text{dyn}}$ the residual of the
estimated parameters on their own learning throws, and $b_{\max}$ a bound on the evidence one motion can
remove, which keeps the term on the scale of Monty's bounded evidence. The term is added to every
hypothesis of every object. The burst sampler then treats it as it treats a feature \eqref{eq:burst}: hypotheses that
the motion contradicts stop growing and are deleted, and if all are contradicted a burst samples new
ones. Second, the motion step transports the hypotheses: later observations are mapped back by
$C^{-1}$, where $C$ is the tracked displacement of the object since the episode began, which is
equivalent to moving every hypothesis by $\Ad_C$. Third, sensing continues, and hypotheses sampled
after a motion are scored against every earlier motion when they are created. The motion step uses
only quantities that a surface tracker provides, and the likelihood is the same for any choice of
tracked point. Algorithm~\ref{alg:monty} places the steps in Monty's loop.

The Newton--Euler equations supply the fixed
rigid-body mechanics assumed here, and the transport of twists, wrenches and inertia by the adjoint
follows from representing pose in $\SE$. Attaching an inferred
spatial inertia to each stored object, inferring it with a particle filter, and adding the clipped
likelihood \eqref{eq:dynevidence} to Monty's evidence are modelling choices. Monty's evidence is bounded
and unnormalized, so the sum ranks hypotheses and is not a calibrated posterior over identity, pose and
inertia. The motion evidence acts only on the hypotheses Monty holds or samples, so the results of E1
depend on its burst sampler and on the threshold of its possible set. The assignment of these operations to
neural populations (Sec.~\ref{sec:neural}) is a hypothesis.

\newcommand{\addl}[1]{\textcolor{orange!70!black}{#1}}
\begin{center}\small
\fbox{\begin{minipage}{0.94\linewidth}
\begin{singlespace}
\refstepcounter{algo}\label{alg:monty}\textbf{Algorithm \thealgo}\quad Monty inference with the
dynamical completion. Added steps are marked \addl{\textbf{+}}.\\[3pt]
\textbf{Input:} object models (Sec.~\ref{sec:static}); \addl{\textbf{+} parameters $\theta_\obj$ of each object
\eqref{eq:paramfilter}.}\\[2pt]
1. Sense the first patch; create hypotheses $(\obj,R,\ell)$ with evidence $e$.\\
2. \textbf{repeat} until one hypothesis dominates or the step budget is spent:\\
\hspace*{1.5em} a. move the sensor by $d$; \addl{\textbf{+} map the observation back by $C^{-1}$;}\\
\hspace*{1.5em} b. for each hypothesis, $\ell\leftarrow\ell+R^{\top}d$ and update $e$ by \eqref{eq:evidence};\\
\hspace*{1.5em} c. vote with the other modules;\\
\hspace*{1.5em} d. delete stagnating hypotheses; sample new ones if the best stops growing (burst);\\
\hspace*{1.5em} \addl{e. \textbf{+} score each new hypothesis against every earlier motion \eqref{eq:dynevidence};}\\
\hspace*{1.5em} \addl{f. \textbf{+} if the object moves: add \eqref{eq:dynevidence} to every hypothesis; update $C$.}\\
3. Output the object and its pose \addl{\textbf{+} with its spatial inertia $\mathcal G$; predict motion at any pose with $\Ad_T$.}\\[2pt]
If nothing moves, steps 2e--f never run and Monty's output is unchanged.
\end{singlespace}
\end{minipage}}
\end{center}

\section{Environment and objects}
\label{sec:setup}
All checks and experiments run in simulation, with physics simulated by MuJoCo \citep{mujoco}, a rigid-body simulator widely used in robotics; the law that the model integrates \eqref{eq:law} is a separate
implementation, and the model receives the simulator's state only through what it senses. In C1 the
agent is Monty's embodied agent: a depth camera that views the object through a small patch
($64\times64$ pixels) and is moved over the object by Monty's motor policy. In C2--C4, E1 and E2, to keep
the tests fast, the patch is replaced by point samples of the object's surface, each carrying the
features the patch would report. A throw releases the object at a height of $1.2$~m with a linear
velocity drawn uniformly from $[1,3]\times[-1,1]\times[2,4]$~m\,s$^{-1}$ and an angular velocity drawn
uniformly from $[-8,8]^3$~rad\,s$^{-1}$; the object flies for $0.9$~s in vacuum, with no floor, and its
pose is recorded at $200$~Hz. The \emph{landing error} of a prediction is the distance between the
predicted and the observed position of the object's frame origin at the end of the flight. A tracker
reports the object's pose during the throw: the simulator's pose with optional noise or, in one
repetition, a pose estimated from rendered depth images (Appendix~\ref{sec:tracking}). Accuracies are
reported with Wilson $95\%$ intervals and compared between systems on the same trials with an exact
McNemar test; means over objects are reported with percentile bootstrap intervals.

\begin{figure}[t]\centering
\includegraphics[width=\linewidth]{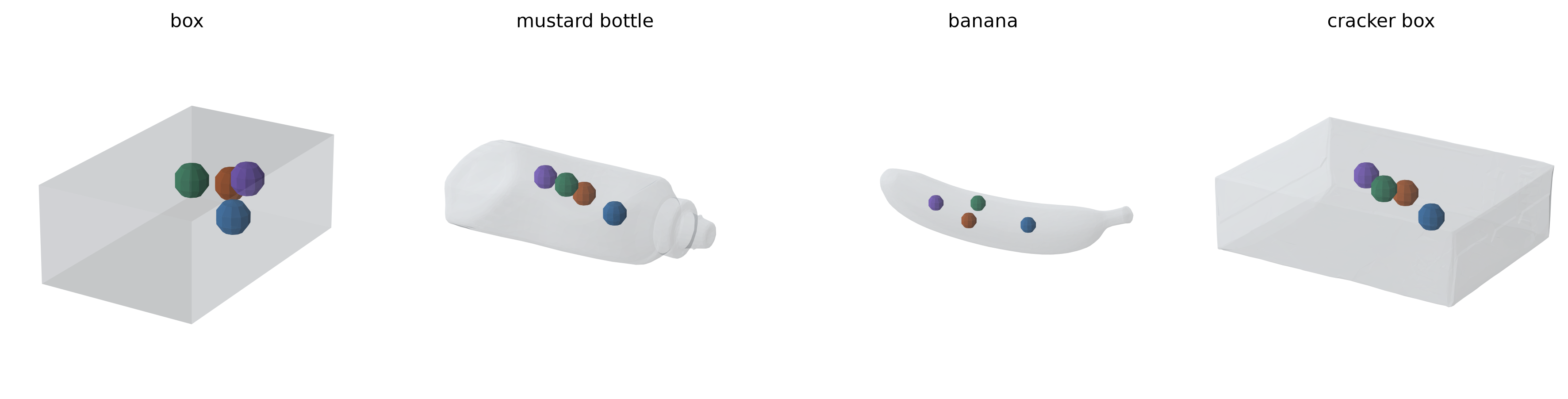}
\caption{Look-alike objects. In each family the four variants share the surface shown and differ in
the position of a hidden mass (coloured spheres, one per variant), invisible to surface sensing.}\label{fig:families}
\end{figure}

\begin{figure}[t]\centering
\includegraphics[width=\linewidth]{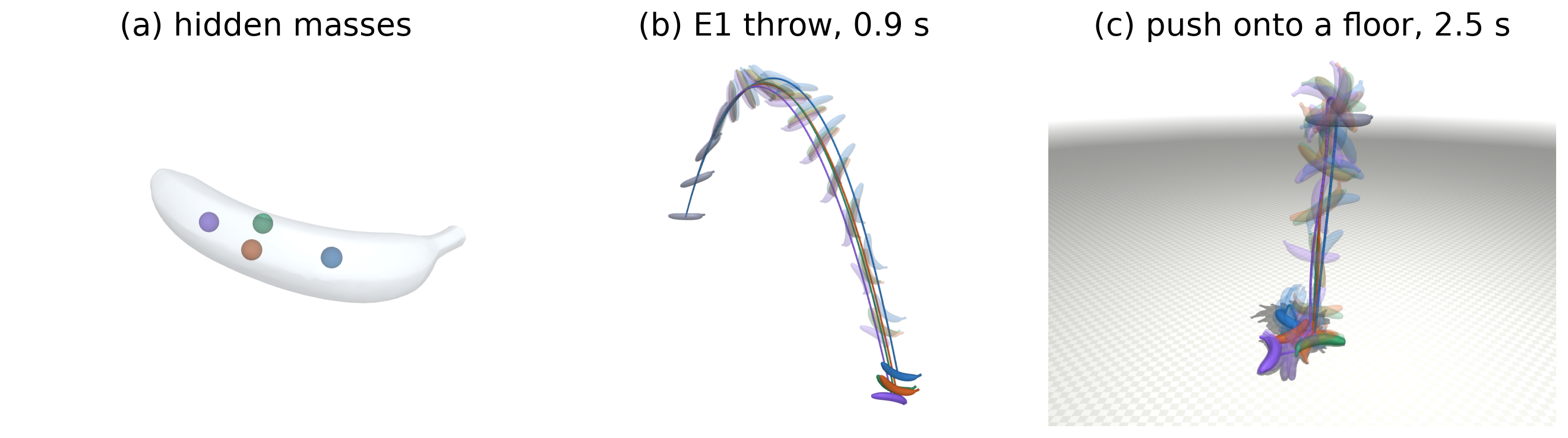}
\caption{The banana family in motion. (a) The four variants; coloured spheres mark the hidden masses.
(b) One throw as in E1: the four variants from the same launch, in free flight for $0.9$~s, with the
trace of each variant's frame origin and snapshots every $60$~ms (colours as in (a)). (c) The same
variants given the same push at a random point of the surface and observed for $2.5$~s on a floor, a
design not used in the experiments. Each panel shows, of $20$ throws, the one whose spread of the
variants' end points is the median ($7$~cm in (b), $15$~cm in (c)).}\label{fig:throws}
\end{figure}

The \emph{look-alike boxes} (Table~\ref{tab:objects}), the objects of E1 and E2, are four boxes with identical surfaces ($10\times16\times6$~cm, density $500$~kg\,m$^{-3}$) that
differ only in a hidden internal mass: a dense sphere whose radius is $35\%$ of the smallest
half-extent and which carries $30\%$ of the total mass ($15\%$ in one condition), centred at $56\%$ of
the half-extent along the box's $x$, $y$ or $z$ axis, or at $39\%$ along all three. The mass is
invisible to surface sensing but displaces the centre of mass and changes the rotational inertia, so
the four boxes look the same and move differently; no $180^\circ$ rotation about a box axis maps one
mass position onto another. The \emph{look-alike YCB families} repeat this design on seven
three-dimensional scans of household items from the Yale--CMU--Berkeley (YCB) object set, a standard
benchmark in robot manipulation \citep{calli}: mustard bottle, banana, cracker box, bleach bottle,
hammer, wood block and the toy drill of the toy-airplane kit, each in four variants whose hidden mass
lies along one of the three principal axes or along a diagonal (Figs.~\ref{fig:families}
and~\ref{fig:throws}). On these shapes the hidden mass moves the centre of mass by $0.2$ to $2.0$~cm,
against $0.5$ to $1.3$~cm in the boxes. The checks C2--C4 use six primitive shapes with hidden masses
and twelve YCB objects of uniform density, whose inertia is MuJoCo's estimate from the mesh (it exceeds
the exact value by $4$--$29\%$ for non-convex meshes; all systems compared observe the same simulated
throws). C1
uses Monty's own benchmarks of ten and $77$ YCB objects.
\begin{table}[h]\centering\small
\caption{Objects. Sizes are full extents.}
\label{tab:objects}
\begin{tabular}{>{\raggedright\arraybackslash}p{2.6cm}>{\raggedright\arraybackslash}p{7.6cm}>{\raggedright\arraybackslash}p{2.6cm}}
\toprule
set & objects & used in \\
\midrule
look-alike boxes & four boxes of $10\times16\times6$~cm with a hidden mass of $30\%$ (or $15\%$) at four positions & E1, E2 \\
look-alike YCB families & mustard bottle, banana, cracker box; four variants of each with a hidden mass of $30\%$ & E1, E2 \\
 & bleach bottle, hammer, wood block, toy drill; as above & E1 \\
primitives & boxes of $10\times16\times6$ and $8\times8\times20$~cm; sphere of diameter $10$~cm; ellipsoid of $14\times9\times6$~cm; cylinders of diameter $8$ and $11$~cm and height $12$ and $4$~cm; each with a hidden mass of $25$--$40\%$ & C2--C4 \\
YCB, uniform density & master chef can, cracker box, sugar box, tomato soup can, mustard bottle, banana, apple, orange, bleach cleanser, bowl, mug, power drill & C2--C4 \\
YCB, Monty's benchmarks & ten distinct objects (mug, bowl, potted meat can, spoon, strawberry, mustard bottle, dice, golf ball, toy brick, banana) and the $77$-object set & C1 \\
\bottomrule
\end{tabular}
\end{table}

\section{The completion preserves recognition and learns object dynamics}
\label{sec:validation}
Four checks test the first question: whether the completion leaves Monty's recognition unchanged (C1),
and whether its law learns an object's dynamics from a few throws (C2), predicts at orientations it has
not seen (C3) and adapts to hidden masses (C4).

\subsection{Monty's recognition is unchanged (check C1)}
\label{sec:q0}
We ran Monty's recognition benchmarks in a port of its depth-camera pipeline to MuJoCo, composed from
Monty's shipped configurations with only the simulator changed, once with \texttt{EvidenceGraphLM} and
once with the completion's subclass (Table~\ref{tab:q0}). With known rotations every output of every
episode is identical, on the ten-object and the $77$-object sets. With random rotations and sensor noise
the pipeline is not deterministic across runs; there the completion's results lie within the
variation of repeated runs of unmodified Monty. The port's accuracy on the $77$-object set is below
Monty's published $98.3\%$ \citep{tbpmonty}, equally in both arms. With two learning modules, voting
ran unchanged with the completion, and the hypothesis-testing policy issued the same targets as with
\texttt{EvidenceGraphLM}.
\begin{table}[h]\centering\small
\caption{C1: recognition in the MuJoCo port of Monty's benchmarks, with Monty's learning module and
with the completion's subclass (episodes correct).}
\label{tab:q0}
\begin{tabular}{>{\raggedright\arraybackslash}p{4.3cm}>{\centering\arraybackslash}p{2.4cm}c>{\raggedright\arraybackslash}p{2.2cm}}
\toprule
benchmark (episodes) & Monty & completion & outputs \\
\midrule
ten objects, known rotations ($140$) & $140$ & $140$ & identical \\
ten objects, random rotations, noise ($100$) & $98$ & $98$ & same identities \\
$77$ objects, known rotations ($231$) & $220$ & $220$ & identical \\
$77$ objects, random rotations, noise ($231$) & $202$, $201$, $201$$^{a}$ & $201$ & within variation \\
\bottomrule
\end{tabular}

\smallskip
\footnotesize $^{a}$ Three runs of unmodified Monty; with sensor noise the pipeline is not deterministic.
\end{table}

\subsection{The law learns, transfers across pose and adapts to hidden masses (checks C2--C4)}
\label{sec:lawchecks}
The parameter filter, started from a prior read off the recognized model, assimilates $k$ observed
throws and predicts held-out throws; we call this estimator the \emph{law arm}, an \emph{arm} being one
of the systems compared. It is compared with two learned predictors (Table~\ref{tab:validation}): a
\emph{regression} fitted to each object's throws, a Gaussian process \citep{sklearn} on the residual of
the ballistic flight of a point mass, and a \emph{network}, an attentive neural process
\citep{garnelo2018} meta-trained on $4000$ simulated objects.

In C2 the arms learn from $k=0$ to $8$ throws. On the primitives the law arm's landing error falls below
$3$~cm after $2.5$ throws on average ($95\%$ interval $1.7$--$3.3$) and reaches $1.1$~cm after eight,
while the regression remains at $3.1$~cm. Before any throw
the law's error ($5.0$~cm) comes from the hidden mass alone: with the true centre of mass it is
$0.4$~cm. On the uniform YCB objects, whose flight is close to ballistic, the regression matches the
law; the same ordering holds with trajectories tracked from rendered depth images
(Appendix~\ref{sec:tracking}). In C3 the arms learn at one orientation and predict at
orientations up to $180^\circ$ away: the law predicts with the same error at all of them, while the
regression's error doubles; regressions without the ballistic term degrade by an order of magnitude
(Appendix~\ref{sec:learnedapp}). In C4 the network learns faster than the law when its
training objects resemble the test objects ($1.5$ against $2.5$ throws) and matches it on new YCB
shapes, but when the hidden masses of the test objects lie outside its training distribution its error
stays at $4.9$--$5.2$~cm from zero to eight throws, while the law arm reaches $1.1$~cm. These checks concern the law alone, so a physics module attached to the recognizer from outside would pass them equally; E1 and E2 test what sharing the hypotheses adds.
\begin{table}[h]\centering\small
\caption{C2--C4: held-out landing error (cm) of the law arm and of the learned comparisons,
on the primitives unless stated; $k$ is the number of observed throws.}
\label{tab:validation}
\begin{tabular}{llcc}
\toprule
check & predictor & error & note \\
\midrule
C2 ($k=1$ / $8$) & law & $3.8$ / $1.1$ & $3$~cm after $2.5$ throws \\
 & regression & $6.5$ / $3.1$ & $3$~cm not reached \\
C2, YCB ($k=0$ / $8$) & law & $0.9$ / $0.7$ & \\
 & regression & $0.8$ / $0.5$ & \\
C3 ($0^\circ$ / $180^\circ$) & law & $1.1$ / $1.1$ & YCB: $0.7$ / $0.7$ \\
 & regression & $2.6$ / $5.0$ & YCB: $0.5$ / $0.8$ \\
C4 in distribution ($k=0$ / $8$) & network & $5.0$ / $1.3$--$1.5$ & $3$~cm after $1.5$ throws \\
C4 masses outside training & network (3 seeds) & $4.9$--$5.2$ / $5.0$--$5.2$ & three training sets \\
 & law & $5.1$ / $1.1$ & \\
C4 new shapes, YCB & network & $1.1$--$1.3$ / $0.7$--$1.1$ & law: $0.9$ / $0.7$ \\
\bottomrule
\end{tabular}
\end{table}

\section{Experiment 1: recognizing look-alike objects from their motion}
\label{sec:e3}
E1 tests the second question: whether, among objects that look the same, motion that enters the
recognizer's hypotheses improves recognition, compared with the same law attached to the recognizer
from outside. Monty learns the four look-alike boxes, with their labels, from the same surface sample, so their
stored models are identical, and each object's parameters are estimated from eight throws. The filter
represents the principal axes of inertia as aligned with the box's axes; the diagonal hidden mass tilts
them by $9^\circ$, which affects all arms equally. In a trial one object is presented at a random
orientation and sensed at $40$ surface points with $2$~mm noise. It is then thrown three times, each
throw starting in the orientation in which the previous one landed, and sensed at $20$ points after
each throw. Identity is scored against the presented object, and pose against its current pose
(geodesic error below $20^\circ$, up to rotations that leave the whole object, hidden mass included,
unchanged).

Three arms see the same trials. \emph{Unmodified Monty} keeps sensing without using the motion. The
\emph{attached module} uses the same law but reads the recognizer from outside: after sensing, it takes
Monty's possible hypotheses, adds every hypothesis that the stored models cannot separate (symmetric
poses, and the same poses of objects with identical models), and re-ranks them by the summed likelihood
\eqref{eq:dynevidence} of the throws. The \emph{completion} runs the motion step inside Monty
(Sec.~\ref{sec:attach}). Simpler attached modules, which follow Monty's most likely hypothesis or
re-rank its possible set without the added hypotheses, identify fewer objects overall
(Appendix~\ref{sec:e3app}).

\begin{figure}[h]\centering
\includegraphics[width=\linewidth]{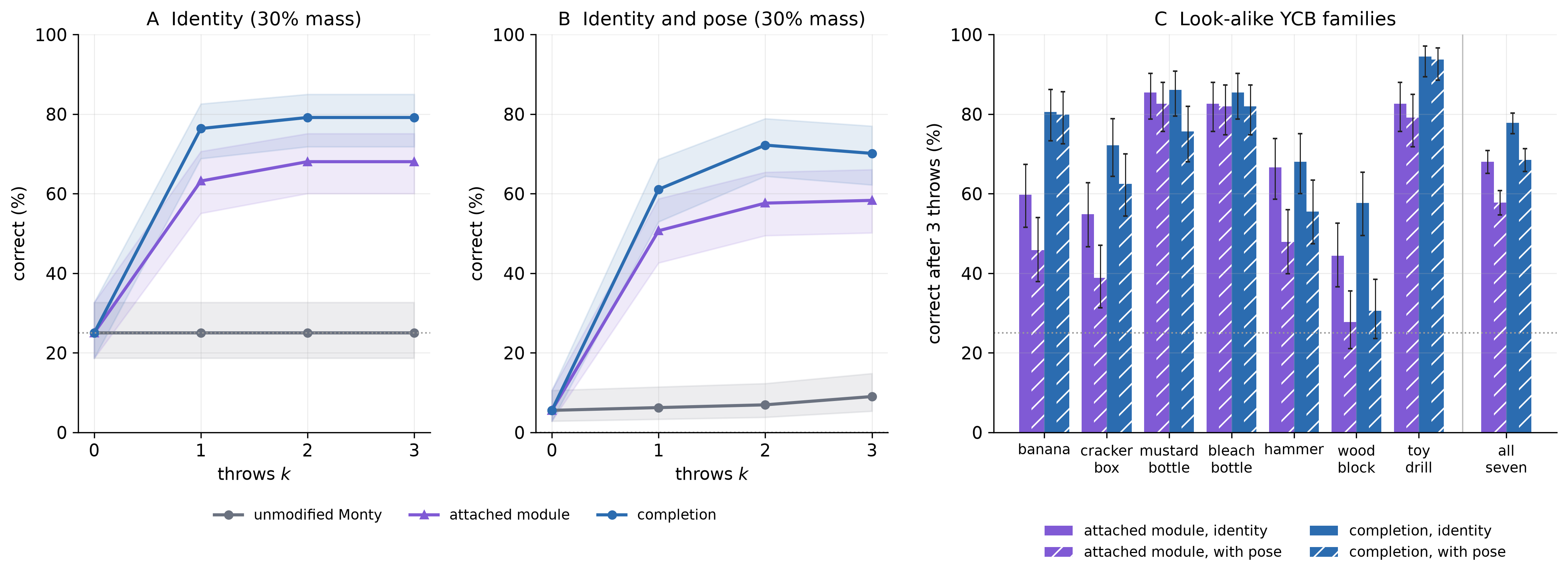}
\caption{E1, recognition from motion. (A) Identification of four objects with identical surfaces
against the number of throws (hidden mass $30\%$, no tracking noise, $144$ trials); bands, $95\%$
Wilson intervals; dotted line, chance. (B) Identity and pose both correct. (C) Look-alike YCB
families: identification (solid) and identification with pose (hatched) after three throws for the
attached module and the completion, $144$ trials per family; bars, $95\%$ Wilson intervals.}\label{fig:e3}
\end{figure}

Surface sensing leaves identification at chance ($25\%$; Table~\ref{tab:e3}, Fig.~\ref{fig:e3}), and
unmodified Monty stays there. After three throws the attached module identifies $68\%$ of the objects
and the completion $79\%$; with pose, $58\%$ and $70\%$. Against the attached module the completion is right where the module is
wrong in $28$ trials and the reverse in $12$ (McNemar, $p=0.017$); with pose, $33$ and $16$
($p=0.021$). Its prediction of held-out throws falls to $2.9$~cm, against $3.3$~cm for the attached
module and $1.0$~cm with the true object and pose. This is because an attached module can only choose among the hypotheses it reads, and Monty's possible set before the throws often lacks the true object and pose,
which the surface cannot separate. In the completion the motion evidence enters every hypothesis Monty
holds or samples later, and sensing continues after each throw. The same access can be obtained from
outside: when the attached module re-reads Monty's hypotheses after every throw, it reaches $84\%$ and
$78\%$ on the same trials ($p=0.14$ and $0.07$ against the completion), with about $240$ law rollouts
per trial against $5\times10^4$. Recognition therefore improves when motion evidence is applied to the hypotheses as sensing revises them, which the completion does by construction. With a lighter hidden mass, shorter sensing or tracking
noise, every arm identifies fewer objects and the completion stays ahead (Appendix~\ref{sec:e3app}).

\begin{table}[h]\centering\small
\caption{E1, recognition from motion, hidden mass of $30\%$, no tracking noise ($144$ trials: four objects, six orientations, six
seeds). Identification and identification with pose after $k$ throws, with $95\%$ Wilson intervals
at $k=3$, and held-out landing error after three throws.}
\label{tab:e3}
\begin{tabular}{lccccc}
\toprule
arm & \multicolumn{2}{c}{identity (\%)} & \multicolumn{2}{c}{identity and pose (\%)} & landing \\
 & $k{=}1$ & $k{=}3$ & $k{=}1$ & $k{=}3$ & (cm) \\
\midrule
unmodified Monty & $25$ & $25$ ($19$--$33$) & $6$ & $9$ ($5$--$15$) & -- \\
attached module & $63$ & $68$ ($60$--$75$) & $51$ & $58$ ($50$--$66$) & $3.3$ \\
completion & $76$ & $79$ ($72$--$85$) & $61$ & $70$ ($62$--$77$) & $2.9$ \\
true object and pose & & & & & $1.0$ \\
\bottomrule
\end{tabular}
\end{table}

The same protocol on the seven look-alike YCB families ($144$ trials per family: four variants, six
orientations, six seeds) gives the same result (Fig.~\ref{fig:e3}C). The completion identifies the object in $78\%$ of
$1008$ trials and the object with its pose in $69\%$, against $68\%$ and $58\%$ for the attached module
(Table~\ref{tab:e3ycb}; discordant trials $197$ against $99$, $p=1\times10^{-8}$, and $230$ against
$121$, $p=6\times10^{-9}$). The gain is significant on the banana, the cracker box, the toy drill and,
for identity, the wood block. An eighth family, built on the spray bottle, is reported apart: its scan
is an open surface in four disconnected pieces, so its simulated inertia does not correspond to a
closed bottle. On both object sets, then, motion evidence that enters Monty's hypotheses identifies
look-alike objects and their pose more often than the same law reading the hypotheses once. Re-reading
them after every throw, the attached module reaches $77\%$ and $68\%$ on the YCB trials ($p=0.46$ and
$0.95$ against the completion), as it does on the boxes.
\begin{table}[h]\centering\small
\caption{E1 on the look-alike YCB families after three throws: identification / identification with
pose (\%), $144$ trials per family, and McNemar $p$ for the completion against the attached module.
$^{\dagger}$Open, fragmented scan: simulated inertia unreliable; excluded from the pooled row.}
\label{tab:e3ycb}
\begin{tabular}{lccc}
\toprule
family & attached & completion & $p$ \\
\midrule
banana & $60$ / $46$ & $81$ / $80$ & $5\times10^{-6}$ / $5\times10^{-11}$ \\
cracker box & $55$ / $39$ & $72$ / $62$ & $0.003$ / $6\times10^{-5}$ \\
mustard bottle & $85$ / $83$ & $86$ / $76$ & $1.0$ / $0.17$ \\
bleach bottle & $83$ / $82$ & $85$ / $82$ & $0.61$ / $1.0$ \\
hammer & $67$ / $48$ & $68$ / $56$ & $0.86$ / $0.17$ \\
wood block & $44$ / $28$ & $58$ / $31$ & $0.02$ / $0.68$ \\
toy drill & $83$ / $79$ & $94$ / $94$ & $9\times10^{-4}$ / $2\times10^{-4}$ \\
all seven & $68$ / $58$ & $78$ / $69$ & $1\times10^{-8}$ / $6\times10^{-9}$ \\
\midrule
spray bottle$^{\dagger}$ & $74$ / $69$ & $81$ / $77$ & $0.15$ / $0.17$ \\
\bottomrule
\end{tabular}
\end{table}

\section{Experiment 2: discovering look-alike objects from their motion}
\label{sec:e5}
E2 tests the third question: whether motion reveals, without labels, that one surface covers several
objects. The four look-alike boxes arrive unlabelled in a stream of $40$ episodes, each a recognition followed
by five throws. Monty stores one surface model, since the objects look the same. The completion
attaches to it a list of physical variants, each a particle filter over the parameters; for each
episode it compares the marginal likelihood of the throws under every existing variant and under a new
one, with a Chinese-restaurant-process prior (concentration $\alpha=1$), and the episode joins the most
probable. Because early variants are broad, the stream is followed by three sweeps that propose to
split each variant in two and to merge pairs of variants, accept a proposal when the marginal
likelihood plus the log prior of the partition increases, and reassign episodes
(Appendix~\ref{sec:splitapp}). The logic is that of contextual inference in motor learning, which
creates a new context when no stored one explains the dynamics \citep{heald2021, wolpert1998}, applied
here to objects. The task makes each hidden-mass configuration a stable identity by construction: the
object of an episode does not change, and the variants are physical variants attached to one surface
model.

In every seed the completion ends with four variants, each holding the ten episodes of one object, so
that its partition of the training episodes equals the true identities (Table~\ref{tab:e5}). The online
pass finds the number of objects but mixes their membership (adjusted Rand index $0.29$--$0.58$),
because an episode can join a variant while it is still broad; the sweeps correct the partition. The
sweeps revisit stored episodes after the stream, an offline step that may correspond to consolidation;
online, the completion finds the number of objects and only part of their membership. On
new episodes the discovered variants identify the object in every case and predict a held-out throw
with $0.64$~cm error, as well as variants learned from labels ($98\%$, $0.85$~cm), while one variant
learned from all episodes predicts with $4.37$~cm.
\begin{table}[h]\centering\small
\caption{E2, discovery from motion: four objects that share one surface (means over three seeds; $40$ training and
$20$ test episodes per seed). The adjusted Rand index compares the assignment of training episodes with
the true identities ($1$: identical partitions, $0$: chance), after the online pass and after the sweeps.}
\label{tab:e5}
\begin{tabular}{lccc}
\toprule
& discovered variants & one merged variant & variants from labels \\
\midrule
variants (true number $4$) & $4$, $4$, $4$ & $1$ & $4$ \\
adjusted Rand index & $0.42\to1.00$ & -- & -- \\
test identification (\%) & $100$ & -- & $98$ \\
test landing error (cm) & $0.64$ & $4.37$ & $0.85$ \\
\bottomrule
\end{tabular}
\end{table}

On the look-alike YCB families discovery fails: the mustard-bottle stream ends with nine variants
(adjusted Rand index $0.19$). The throws carry too little information to separate the variants:
along the shortest axis of these shapes the hidden mass moves the centre of mass by $0.2$ to $0.5$~cm,
about the precision of one episode ($\pm0.2$~cm), and the surface sample leaves the pose uncertain by
a median of $10^\circ$, against under $1^\circ$ for the box. Even variants learned from labels
identify only $55\%$ of the mustard-bottle test episodes. Where one episode cannot separate two
configurations, its marginal likelihood does not distinguish a new variant from noise, and the stream
creates variants without a physical difference behind them. This is a limit of the current procedure
under the tested throws and stream length; evidence that is weak in each episode can accumulate over
more episodes or more informative interactions.

\section{A candidate neural implementation and testable predictions}
\label{sec:neural}
This section asks whether the completion's central operation can be computed by a population of
neuron-like units, and what such an implementation would predict; it does not claim that cortex uses
this implementation. In the laminar account of the Thousand Brains Theory, layer 6a codes the sensor's
location in the object's frame and the orientation of the object relative to the sensor is carried
alongside \citep{lewis2019, tbt2019}. The second version of the theory adds the operation that turns
sensor movement into movement on the object: thalamic relay cells rotate incoming movement signals and
sensed features from the sensor's orientation into the object's, using the orientation inferred in
layer 6b \citep{tbt2025}. The completion adds to these transformations the quantities that the law of
motion needs, the twist and the spatial inertia, and the consequence that observed motion is evidence
for identity and pose. Frontoparietal regions in
humans respond selectively during physical prediction and carry object mass and stability in a form
that generalizes across scenes \citep{fischer2016, schwettmann2019, pramod2022}. In human superior
parietal lobule, units recorded with microelectrode arrays prefer physical over control conditions in two
different tasks, and the same population predicts whether a moving object will collide about 400~ms
before contact, in a form that generalizes across scenarios \citep{bougou2026}.

\subsection{A gain-field network for the velocity transport}
\label{sec:neuraladjoint}
We built a minimal network for the rotational part of the adjoint, $\omega_b=R^{\top}\omega_w$, which
expresses a world-frame angular velocity $\omega_w$ in the object's frame given its recognized
orientation $R$. This is the rotation the second version of the theory assigns to thalamic relay cells.
There, each relay cell receives several inputs and passes on one of them, selected by the orientation
signal, as a multiplexer does \citep{tbt2025}. A gain field computes the same rotation by multiplying a
velocity code with an orientation code, so its output varies smoothly with orientation
\citep{zipser1988, pouget1997}. Our network follows the transformation circuit of spatial-memory models,
in which head-direction cells gate the rotation of egocentric boundary vectors into allocentric ones
\citep{byrne2007, bicanski2018}, with the object's recognized orientation in place of heading. Forty-two input
units have cosine tuning to $\omega_w$ around preferred axes $u_i$
spread over the sphere; $128$ input units are tuned to orientation with an antipodally symmetric
(Bingham-like) profile around preferred orientations $R_j$ of half-width set by a concentration
$\kappa$; both emit Poisson counts in $0.5$~s. An intermediate layer multiplies each velocity unit by
each pool-normalized orientation unit, as in basis-function models of object-centred coding \citep{denevepouget2003}, and $42$
output units with preferred body axes $b_k$ read the layer linearly. A readout fixed by the geometry,
$W_{k,ij}=\rho\,b_k^{\top}R_j^{\top}u_i$ with only a scale $\rho$ fitted, computes the transport to within
$10.9^\circ$ (median angular error, against $8.7^\circ$ for an observer that knows $R$ exactly), both at
orientations used to fit the scale and in a held-out $75^\circ$ region of orientations. A readout learned
by ridge regression outside that region matches it inside but degrades in the held-out region as the
orientation tuning narrows ($16^\circ$ to $30^\circ$ at a half-width of $17^\circ$;
Fig.~\ref{fig:neuraladjoint}a). Measured as an
experimenter would, the output units' preferred world axes rotate with the object: they match $Rb_k$
with a median error of $3^\circ$, and a $110^\circ$ rotation of the object shifts the tuning peaks of two
example units by $109^\circ$ and $111^\circ$ (Fig.~\ref{fig:neuraladjoint}b,c).

\subsection{Predictions}
Four predictions follow, ordered by how much they assume beyond the model. First, co-rotation:
velocity tuning in an object's frame should rotate with the object's recognized pose, without
re-learning. Object-centred coding is established for static position and discrete direction
\citep{hoydal2019, olson1995, crowe2008}, and motion tuning in egocentric and world frames
\citep{sasaki2020}, but motion tuning anchored to a single object's pose has not been reported; the
second version of the Thousand Brains Theory predicts the static counterpart for oriented edges
\citep{tbt2025}. Co-rotation is compatible with other architectures that use object-centred frames,
so it supports the completion without singling it out. A paradigm that records parietal motion tuning
while objects rotate at trained and novel orientations, adapted from \citet{sasaki2020}, could test it
directly; intracortical arrays in human superior parietal lobule already record physics-preferring units
during physical prediction \citep{bougou2026}.

Second, position-dependent compensation, which distinguishes rigid-body transport from a rotation of
movement signals. For a turning object, the completion adds a term that a rotation alone does not supply. Let the object
move with linear velocity $v_c$ of its centre of mass and angular velocity $\omega$, and let a sensor
patch move with velocity $v_s$ at offset $r$ from the centre of mass, all in world coordinates. The
patch then moves over the object's surface with velocity $R^{\top}\bigl(v_s-v_c-\omega\times r\bigr)$ in the
object's frame: the object's twist, transported to the patch by the adjoint, is subtracted before the
rotation. The law supplies $v_c$ and $\omega$ between observations. The movement signal that updates
the sensor's location on the object should therefore depend on where the patch lies on a turning object,
with magnitude $|\omega\times r|=|\omega|\,|r|\sin\theta$, where $\theta$ is the angle between $r$ and
the rotation axis, so the correction vanishes for a patch on the axis; a rotation of movement signals by
the object's orientation produces no such dependence. In E1's throws the box spins at a median of
$8$~rad/s, so for a patch $4$~cm from the centre of mass and roughly perpendicular to the axis the
correction is about $0.3$~m/s, of the order of the motion of the patch itself. Unlike co-rotation, this
prediction separates rigid-body movement compensation from a rotation of movement signals, and the
measured spin and the patch's position fix the size of the effect; it does not by itself establish that
inertia is inferred or that the completion is organized as proposed here.

\begin{figure}[t]
\centering
\includegraphics[width=\linewidth]{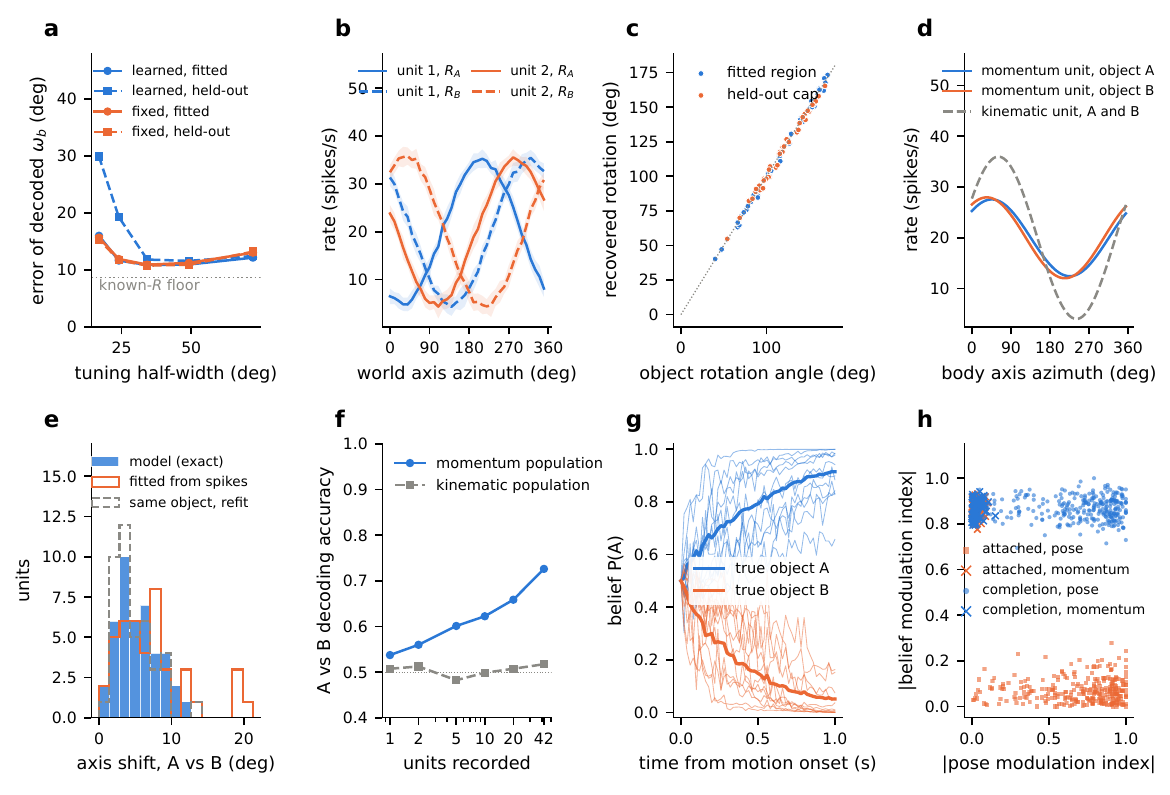}
\caption{A gain-field network for the velocity transport. (a)~Median error of the decoded body-frame
angular velocity against the half-width of the orientation tuning, for learned and fixed readouts, at
fitted and held-out orientations; dotted, an observer that knows the orientation. (b)~Tuning of two
output units to the world-frame rotation axis at two object orientations $110^\circ$ apart. (c)~Object
rotation recovered from the measured tuning against the true rotation. (d)~A momentum unit for two
look-alike objects and an object-agnostic unit. (e)~Shift of preferred axes between the two objects.
(f)~Identification of the object from one population response. (g)~Belief in object A during $1$~s of
tumbling. (h)~Pose and belief modulation of units in the completion and in an attached
module.}\label{fig:neuraladjoint}
\end{figure}

Third, inertia as a gain, which assumes a population code for angular momentum
$L=\mathbf{I}_{\text{c}}\,\omega_b$, with $\mathbf{I}_{\text{c}}$ the rotational inertia about the centre
of mass: the preferred body axis and modulation depth of its units depend on the object's mass
distribution. For two of the look-alike boxes the preferred axis shifts by a median of $4.8^\circ$
between objects, too little to detect in single units, but a population of $42$ units identifies the
object from one $0.5$~s response ($73\%$, against $52\%$ for an object-agnostic code;
Fig.~\ref{fig:neuraladjoint}d--f). The Thousand
Brains Project's proposed behaviour models are object-agnostic \citep{tbpbehaviors}, so they predict
no such dependence.

Fourth, shared hypotheses, which assumes that object hypotheses are held in separate populations: the
same pose-tuned units should change their response to an identical probe motion according to which
look-alike object moved before it. In simulation, $87\%$ of pose-tuned units carried both properties
in the completion and none in an attached module (Fig.~\ref{fig:neuraladjoint}g,h); attached modules
with feedback or mixed selectivity lie in between ($1$--$55\%$ and $38\%$;
Appendix~\ref{sec:neuralrobust}). If pose-tuned units are tuned to the product of the frame's inertia, weighted by the belief over the
two objects, and the angular velocity, the motion evidence also rotates their preferred axes, by about
$3.5^\circ$ in simulation; the attached variants, in which the belief changes only gains at a fixed
pose, do not produce it. Feedback that revises the inferred pose would rotate the preferred axes as well,
so the signature is informative only when the pose is fixed, for example by a probe at a pose already
confirmed by touch; there it distinguishes the inertia-dependent tuning model from the gain-only
alternatives, without by itself locating where the inertia is stored. These results hold across population
sizes, tuning widths, counting windows, correlated noise and perturbed readout weights
(Appendix~\ref{sec:neuralrobust}).

\section{Discussion}
\label{sec:discussion}

\subsection{Relation to other work}
\label{sec:related}
The Thousand Brains Project proposes that object behaviours be modelled as sequences of changes at
locations, in frames and columns of their own, object-agnostic and linked to objects by hierarchy
\citep{tbpbehaviors}. That proposal is kinematic, describing change without forces or mass, and to our knowledge it has not yet been implemented. The two designs fit different parts of rigid-body motion: what all rigid bodies share, the law and the adjoint, is
object-agnostic in the completion as well; what differs between them, the spatial inertia, is a
property of one object expressed in its frame, and E1 and E2 show what is gained by keeping that
property in the object's own hypotheses. Learned behaviour models could cover articulated and
deformable changes, which lie outside the rigid-body case.

Simulation accounts of intuitive physics infer object properties by running a physics engine
\citep{battaglia2013, hamrick2016, ullman2018, smith2019adept, smith2024}; they have been criticized
because people's judgements show conjunction fallacies and task-specific errors \citep{ludwinpeery2020,
ludwinpeery2021, smith2018tasks, kubricht2017, bass2021}. In these accounts, and in analysis-by-synthesis
models that infer mass from video \citep{wu2015galileo, wu2016physics101, sanborn2013}, objects are
given or detected upstream. Visual de-animation trains a perception network through a physics engine,
but at inference the perceived state does not receive evidence from motion \citep{wu2017deanimation};
physics has been used to reject physically unstable pose hypotheses in clutter \citep{mitash2018}, by
testing static consistency between objects. Joint estimation of a body's
pose, velocity and inertia ratios from its motion is established in space robotics \citep{tweddle2015},
and robots recognize objects by interacting with them \citep{bohg2017, sinapov2011}; these systems do
not maintain hypotheses about identity and pose that the motion revises. Learned world models adapt to
new dynamics through context vectors \citep{garnelo2018, nagabandi2019, kumar2021rma}, and video models
acquire some intuitive physics \citep{assran2025, garrido2025} but remain near chance at inferring
latent properties such as mass \citep{tung2023}. Geometric-algebra transformers learn object dynamics with a soft bias
toward rigid motions, which helps where walls and contacts break symmetry \citep{linander2026soft};
the dynamics are learned end to end from object states, without latent physical parameters or
evidence returned to recognition. Group-theoretic accounts derive grid codes from the
requirement that self-motion acts as a group representation \citep{gao2021}; the completion asks the
same of an object's frame. The contribution here is a single hypothesis space for identity, pose and
inertia, inside a working reference-frame recognizer, in which motion is evidence. It is one principled
way to obtain these properties, and other organizations can reach similar accuracy, as the attached
module does when it re-reads the recognizer after every throw (Sec.~\ref{sec:e3}).

\subsection{Limitations}
\label{sec:limits}
E1 and E2 run in simulation: objects are sensed as surface point samples, and the motion is the
simulator's pose with optional tracking noise; only C1 uses Monty's depth-camera pipeline. The
experiments fix one design among many: a hidden mass at $39$--$56\%$ of the half-extent carrying
$30\%$ of the mass, a launch that does not depend on the object, and $0.9$~s of free flight. A heavier
mass nearer the surface, a push at a random point of the surface, which gives each variant a different
spin from the same impulse, and a longer observation ending on a floor all separate the variants
further: over $20$ throws of the banana, the median spread of the variants' end points is $7$~cm with
E1's throws and $15$~cm with random pushes onto a floor (Fig.~\ref{fig:throws}). The law's drag and
contact terms were tested separately (Appendix~\ref{sec:strataapp}); E1 and E2 use neither. In E2 an
object's mass distribution does not change; an object whose load shifts while it is tracked, such as a
bottle being emptied, would require separating the object's identity from a changing physical context,
which these experiments do not test. The
predictions about people are untested; a direct test shows observers the throws of E1 as
three-dimensional replays, after the same eight learning throws as the model, so that each response
pairs with the model's on identical input.

\subsection{Conclusion}
\label{sec:conclusion}
The completion can be added to Monty without changing its
recognition, and its law learns an object's dynamics from a few throws, predicts at unseen
orientations and adapts to hidden masses (C1--C4). Sharing the recognizer's hypotheses lets motion
improve recognition: on look-alike boxes and on look-alike versions of seven household objects the
completion identifies the object and its pose more often than the same law attached from outside and
reading the hypotheses once; read after every throw, the attached law reaches similar accuracy, so the
benefit lies in motion evidence applied to the hypotheses as sensing revises them (E1). Without labels, motion reveals online that one surface covers four boxes, and offline refinement recovers each of them,
while on the household shapes the tested throws do not separate the variants (E2). A gain-field network computes the
completion's central operation, the transport of velocity between frames, and predicts velocity tuning that rotates with an object's recognized
pose. The next tests are motion tracked from images and people judging the same throws.

\appendix
\section{Methods}
\label{sec:methods}

\subsection{Parameter filter and priors}
Parameters are held as $z=(\log m,\ r_{1:3},\ \log I_{1:3},\ \log c^d,\ \mathrm{logit}\,\varepsilon)$, with
$I_{1:3}$ the principal inertias, $c^d$ the drag coefficient and $\varepsilon$ the restitution, and
mapped to physical values with floors that keep each particle's spatial inertia well conditioned
($r$ clipped to $\pm0.5$~m, $I$ floored at $10^{-7}$). The filter uses $J=3000$ particles,
$\sigma=0.04$, $w_{\text{rot}}=0.05$, resampling when the effective sample size $1/\sum_j w_j^2$ falls
below $J/2$, and a move step $z_j\mathrel{+}=b_0\gamma^{i}\Sigma^{1/2}\eta_j$ after the $i$-th resampling
($b_0=0.4$, $\gamma=0.7$, $\eta_j$ standard normal, $\Sigma^{1/2}$ the prior standard deviations).
Predictions use the posterior mean. The law arm of C2--C4 uses a prior read off the
recognized model: centre of mass at the centre of the axis-aligned bounding box of the stored nodes
with standard deviation $\lambda_i/\sqrt3$ on axis $i$ ($\lambda_i$ the half-extent, the spread of a
uniform distribution over the box), principal inertias near those of a solid box of that extent
(standard deviation $0.6$ on $\log I_i$), and $1.5$ on $\log m$, which free flight does not determine.
E1 and E2 use the same form with a fixed standard deviation of $2$~cm on the centre of mass.
Table~\ref{tab:priors} compares five priors for the same law on the primitive set. A prior confident
about the centre of mass speeds learning; the parts that come from the recognized shape, the inertia
and the anisotropy of the bounding box, do not, because the hidden masses move the centre of mass by at
most $1.4$~cm, a small fraction of the box in every direction. The bounding-box prior is used because it
is the only one without a parameter set for these objects.
\begin{table}[h]\centering\small
\caption{Five priors for the same law in C2, on the primitives: interactions to $3$~cm ($95\%$ interval over
objects) and landing error after one and eight throws.}
\label{tab:priors}
\begin{tabular}{lccc}
\toprule
prior on the centre of mass & interactions & $k{=}1$ & $k{=}8$ (cm) \\
\midrule
bounding box, $\lambda_i/\sqrt3$ (law arm) & $2.5$ ($1.7$--$3.3$) & $3.83$ & $1.09$ \\
size only, isotropic & $1.8$ ($1.2$--$2.8$) & $3.26$ & $1.07$ \\
fixed $2$~cm, shape-derived inertia & $1.5$ ($1.2$--$1.8$) & $3.03$ & $0.94$ \\
fixed $2$~cm, generic inertia & $1.5$ ($1.2$--$1.8$) & $2.92$ & $0.92$ \\
broad, $5$~cm, generic inertia & $4.3$ ($4.0$--$5.0$) & $9.73$ & $0.96$ \\
\bottomrule
\end{tabular}
\end{table}

\subsection{Recognizer, and details of Experiment 1}
\label{sec:e3app}
In C2--C4, E1 and E2 recognition uses point samples of the object's surface: each sample carries its location, a
right-handed frame from the surface normal and principal-curvature directions (a discrete shape
operator over $16$ neighbours) and two log principal curvatures. The learning module uses informed
initial poses, a maximum match distance of $8$~mm, the burst-sampling updater (nearest neighbours $5$) and Monty's shipped threshold for the possible set (\texttt{x\_percent}, $20\%$). In E1 the hypothesis frame's origin is the translation that best aligns all sensed points with the
stored model under the hypothesis's orientation (nearest-node least squares, iterated from Monty's
location hypothesis). The tracker reports each throw's displacements, launch angular velocity and the
launch velocity of the frame origin; with tracking noise each tracked pose is perturbed by $3$~mm and
$1.5^\circ$. The width $\sigma_{\text{dyn}}$ is the median residual of the four objects' parameters on
their learning throws, and $b_{\max}=10$. If no object passes Monty's threshold, the attached modules
fall back to the most likely hypothesis (this did not occur). Two simpler attached modules were run on
the same trials: one that follows Monty's most likely hypothesis stays at $25\%$ identification and $6\%$
with pose (landing error $6.3$~cm), and one that re-ranks Monty's possible set without the added
hypotheses reaches $58\%$ and $33\%$ ($4.8$~cm) on the boxes and $67\%$ and $55\%$ on the seven YCB
families, against $68\%$ and $58\%$ for the attached module of the main text. Held-out landing error is the mean over
three throws from the presented orientation, predicted with the identified object at the identified
orientation. A trial with the completion takes about $90$~s on one GPU.

With a lighter hidden mass ($15\%$), shorter sensing ($15$ steps) or tracking noise of $3$~mm and
$1.5^\circ$ per frame, every arm identifies fewer objects and the completion stays ahead
(Table~\ref{tab:e3cond}); with $48$ trials per condition the difference is significant only for
identity with short sensing.
\begin{table}[h]\centering\small
\caption{E1 after three throws, by condition: identification / identification with pose (\%), and
McNemar $p$ for the completion against the attached module (identity / pose).}
\label{tab:e3cond}
\begin{tabular}{lcccc}
\toprule
condition & trials & attached & completion & $p$ \\
\midrule
$30\%$ mass & $144$ & $68$ / $58$ & $79$ / $70$ & $0.017$ / $0.021$ \\
$15\%$ mass & $48$ & $54$ / $40$ & $65$ / $48$ & $0.13$ / $0.39$ \\
$30\%$, $15$ sensing steps & $48$ & $48$ / $46$ & $77$ / $65$ & $0.001$ / $0.08$ \\
$30\%$, tracking noise & $48$ & $52$ / $42$ & $60$ / $44$ & $0.48$ / $1.0$ \\
\bottomrule
\end{tabular}
\end{table}

\subsection{Details of Experiment 2}
\label{sec:splitapp}
Variants have $1000$ particles over twelve parameters, the nine above and the orientation of the
principal axes relative to the body axes (prior standard deviation $0.25$~rad), since a mass off the
body's axes tilts them ($9^\circ$ for the diagonal mass). Variants learn and are compared at one
likelihood scale, $\sigma=1.5$~cm. The marginal likelihood of an episode under a variant is estimated by
the filter, each throw scored before it is assimilated, and averaged over the pose hypotheses the
surface leaves open. A split proposal groups a variant's episodes by $2$-means on the absolute fitted
centre of mass (unchanged by the box's symmetric poses); the log prior of a partition into $B$ blocks of
sizes $n_b$ is $B\log\alpha+\sum_b\log\Gamma(n_b)$, with $\Gamma$ the gamma function. Test episodes are assigned to the most likely variant,
and each variant is mapped to the object most of its training episodes came from. Clustering the
centres of mass fitted to each episode alone, with the number of clusters chosen by the Bayesian
information criterion, recovers the box partition in one seed of three (adjusted Rand index $1.00$,
$0.75$, $0.87$) and reaches $0.27$ to $0.74$ on the YCB families.

\subsection{Learned comparisons (checks C2--C4)}
\label{sec:learnedapp}
The regression is a Gaussian process with a constant times radial-basis-function kernel plus white
noise, on the launch twist and the gravity direction in the body frame, fitted to the residual of the
ballistic flight of a point mass. Without the ballistic term, a Gaussian process fitted to the throws in
the world frame remains at $109$ and $19$~cm after one and eight throws on the primitives ($8.3$~cm on
the uniform YCB objects), and in C3 its error grows from $18$ to $262$~cm at $180^\circ$ ($8.5$ to
$384$~cm on YCB); an equivariant version, predicting in the body frame and rotating into the world, grows
from $18$ to $433$~cm ($8.5$ to $483$~cm). The meta-trained network is an attentive neural process: each
context throw (its launch twist, gravity direction, final displacement and twelve poses through the
flight, in the initial body frame) is a token; tokens pass one self-attention layer with a learned null
token, and each target throw attends over them (four heads, width $128$); the output is a Gaussian over
the final displacement in the initial body frame. Training sets have $4000$ objects with $16$ throws
each (boxes, ellipsoids, cylinders; density $500$~kg\,m$^{-3}$), $30\,000$ steps of the AdamW optimizer, $0$--$8$
context throws per example. The in-distribution set draws hidden masses as the test objects do; the
shifted sets have uniform density, hidden masses only within $35\%$ of the half-extent of the centre
($0$--$60\%$ of the mass), or hidden masses only of $0$--$20\%$ of the mass, while every test object's
hidden mass lies at $41$--$53\%$ of the half-extent and carries $25$--$40\%$.

\subsection{Trajectories from images (check C2)}
\label{sec:tracking}
Three cameras render each throw as depth images at $100$~Hz; the stored surface model is registered to
the back-projected points by point-to-plane iterative closest point registration, warm-started from a constant-velocity extrapolation.
Held-out throws start from the tracked launch pose and a launch twist estimated by the tracker or
inferred with the law from the first $0.1$~s; with the true parameters the error is $12.4$ and $8.9$~cm
for the two launch estimates, the floor for every law-based estimate. The law-based estimates stay at
this floor ($9$--$13$~cm), and the regressions without the ballistic term remain at $74$--$103$~cm.

\subsection{Recognition benchmarks (check C1)}
\label{sec:fullsystem}
The recognition benchmarks are four of Monty's shipped configurations, with the simulator changed to
MuJoCo and models pre-trained in the port:
\begin{itemize}[leftmargin=1.5em,itemsep=0pt]
\item \texttt{base\_config\_10distinctobj\_dist\_agent};
\item \texttt{randrot\_noise\_10distinctobj\_dist\_agent};
\item \texttt{base\_77obj\_dist\_agent};
\item \texttt{base\_77obj\_dist\_agent} with random rotations and sensor noise.
\end{itemize}
\noindent the completion enters by setting the learning module's class. Voting ran with two
instances of the completion's module through Monty's model-level voting code, and the hypothesis-testing
goal generator ran with it, with targets and outputs identical to those with \texttt{EvidenceGraphLM}.

\section{Further probes}
These probes characterize the parameter filter and the contact and drag terms of the law; they are not
part of the checks or the experiments.

\subsection{Identifiability}
\label{sec:identifiability}
\begin{figure}[h]\centering
\includegraphics[width=0.85\linewidth]{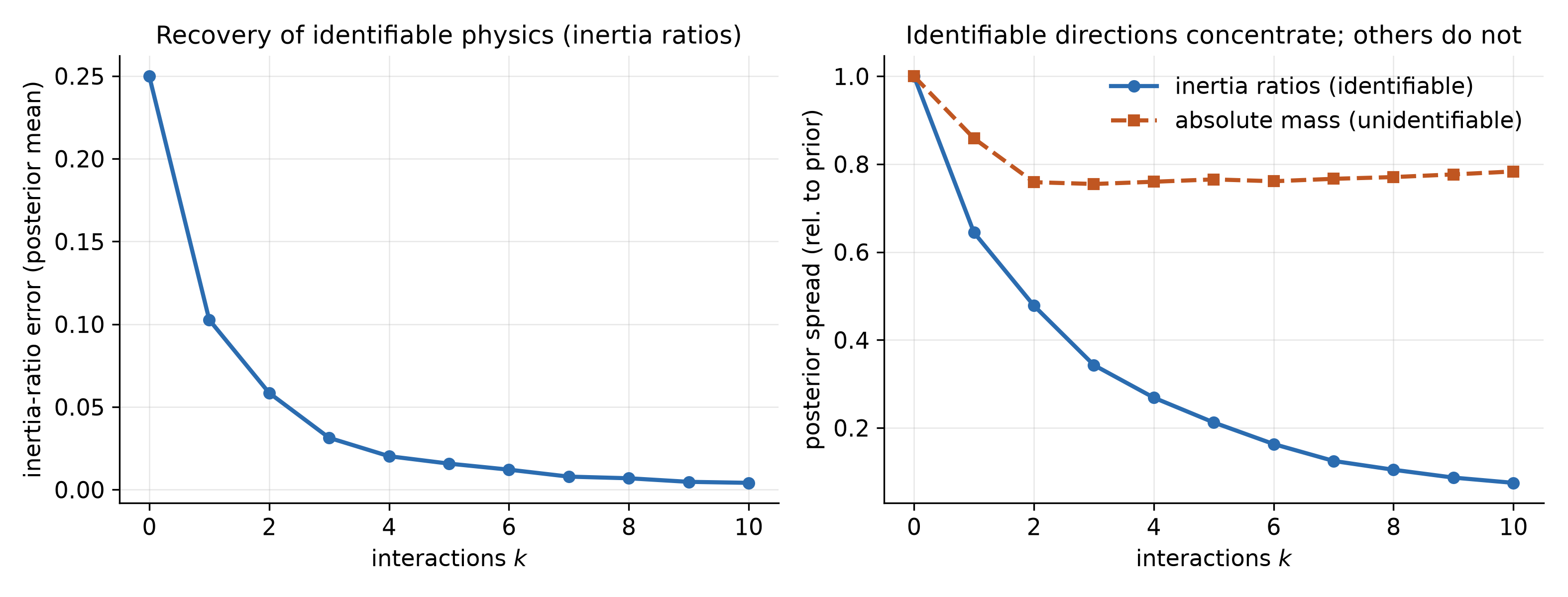}
\caption{Recovery from a broad prior: error of the inertia ratios and posterior spread of inertia
ratios and absolute mass against the number of throws.}\label{fig:recovery}
\end{figure}

\begin{figure}[h]\centering
\includegraphics[width=0.85\linewidth]{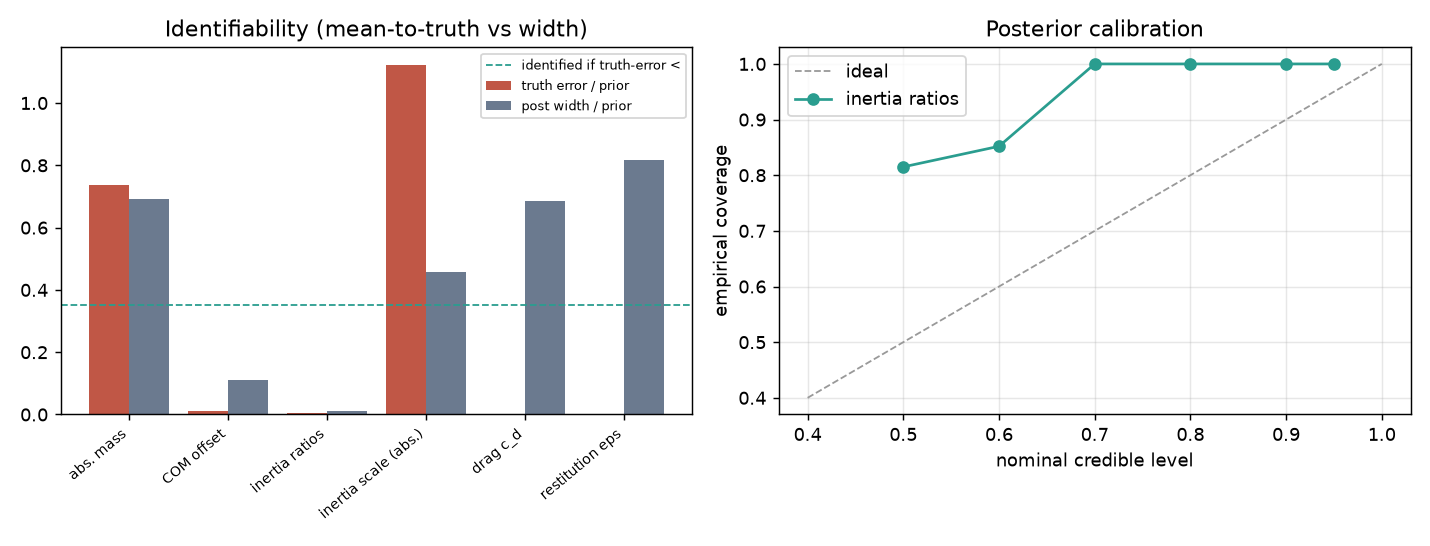}
\caption{Identifiability (error and width of the posterior relative to the prior) and coverage of the
inertia-ratio credible intervals.}\label{fig:calibration}
\end{figure}

From a broad prior on three uniform objects, free flight determines the inertia ratios and the centre of
mass, but not absolute mass or the absolute scale of inertia: after eight throws the error of the
posterior mean, relative to the prior's standard deviation, is $0.006$ for the inertia ratios and $0.01$
for the centre of mass, against $0.74$ and $1.12$ for absolute mass and inertia scale; credible
intervals for the inertia ratios cover at or above their nominal level (Figs.~\ref{fig:recovery},
\ref{fig:calibration}).

\subsection{Contact and drag}
\label{sec:strataapp}
\begin{figure}[h]\centering
\includegraphics[width=0.85\linewidth]{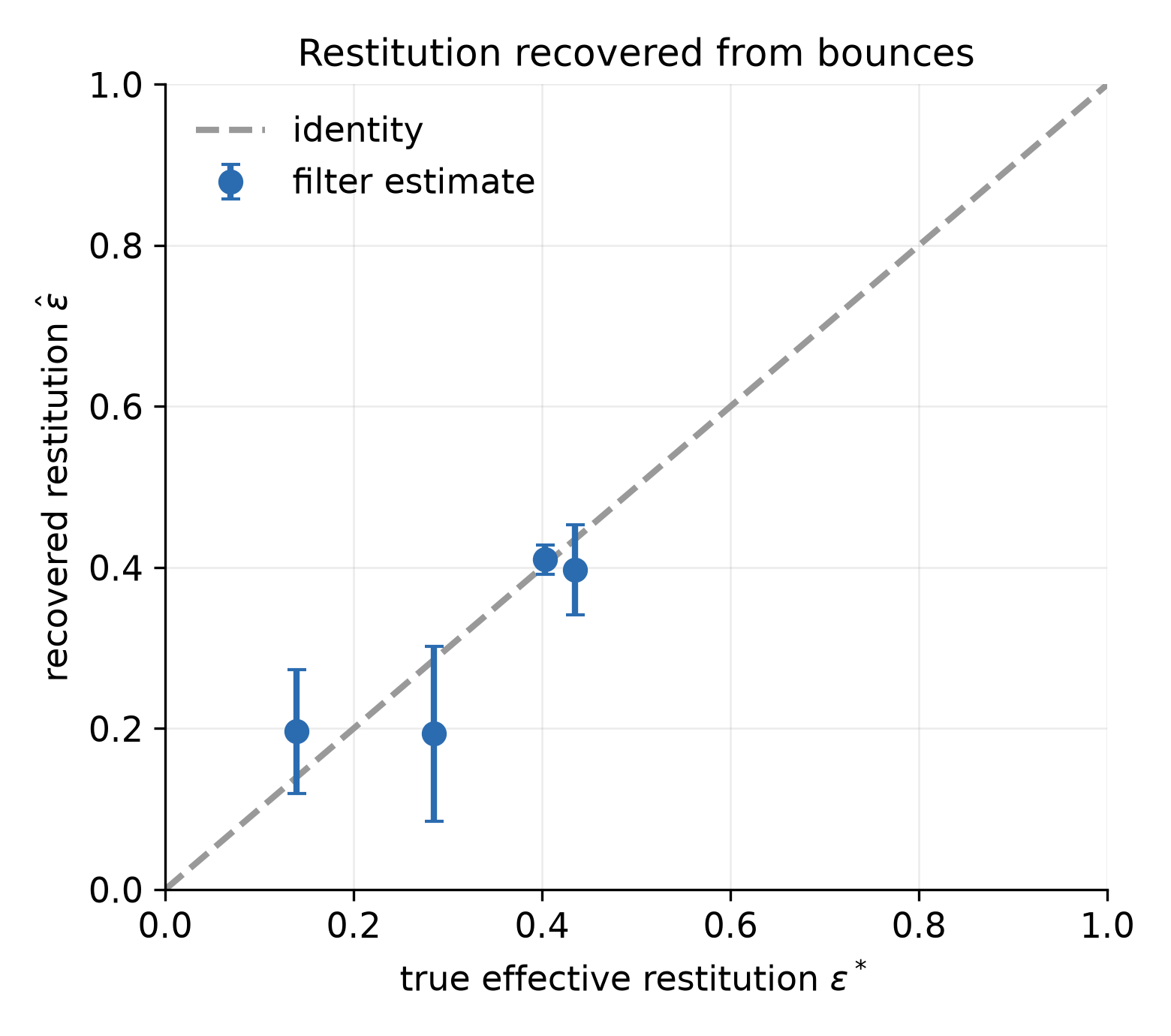}
\caption{Restitution recovered from bounces against each floor's effective restitution.}\label{fig:contact}
\end{figure}

\begin{figure}[h]\centering
\includegraphics[width=0.85\linewidth]{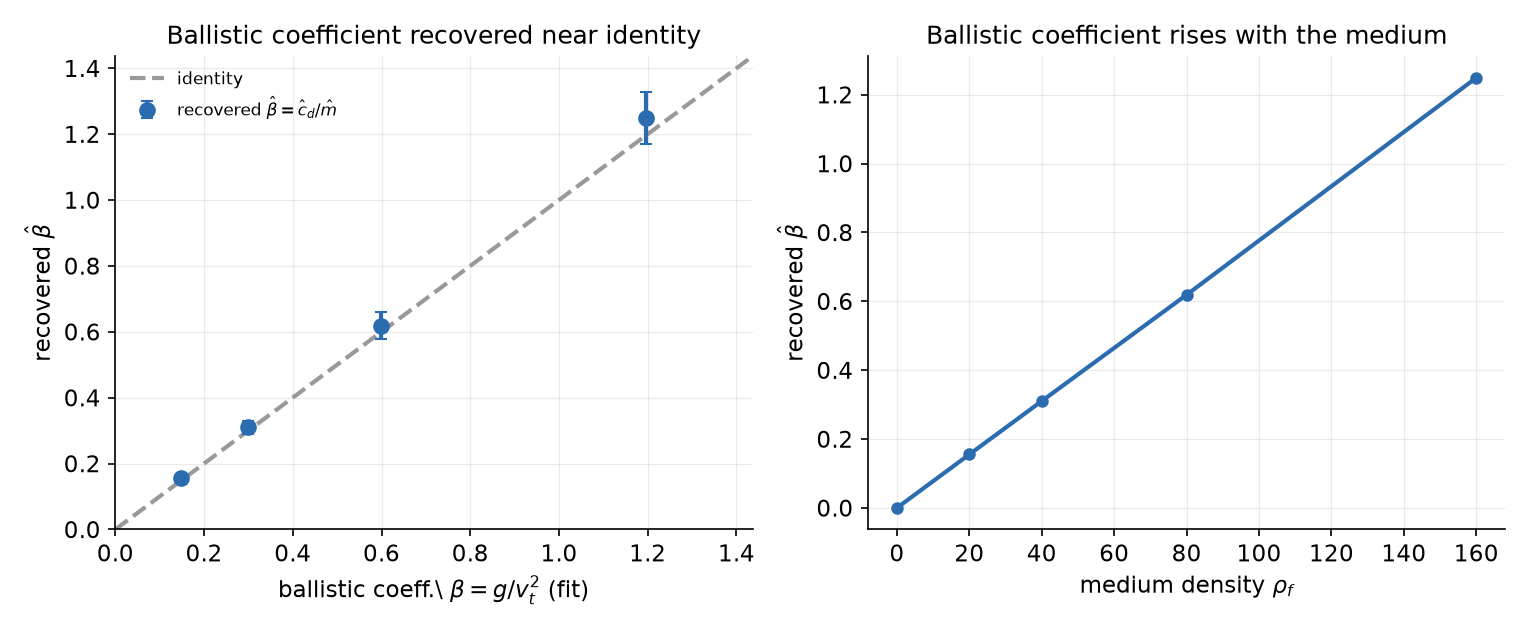}
\caption{Ballistic coefficient recovered from falls through four media against its reference
value.}\label{fig:drag}
\end{figure}

A sphere dropped onto four floors recovers their restitution to within $0.04$ on the two more elastic
floors and to within $0.06$--$0.09$ on the two least elastic, where the simulator's soft contact differs
most from the law's point restitution (Fig.~\ref{fig:contact}). A sphere falling through four media
recovers its ballistic coefficient (drag coefficient over mass) to within $4\%$ on average
(Fig.~\ref{fig:drag}). In a medium the drag term of \eqref{eq:law} is
$\mathcal{F}_{\text{medium}}=-c^d_\obj\lVert v\rVert A\,\mathcal V$ with $A=\mathrm{diag}(0,0,0,1,1,1)$.

\subsection{Robustness of the neural model}
\label{sec:neuralrobust}
\begin{figure}[h]\centering
\includegraphics[width=\linewidth]{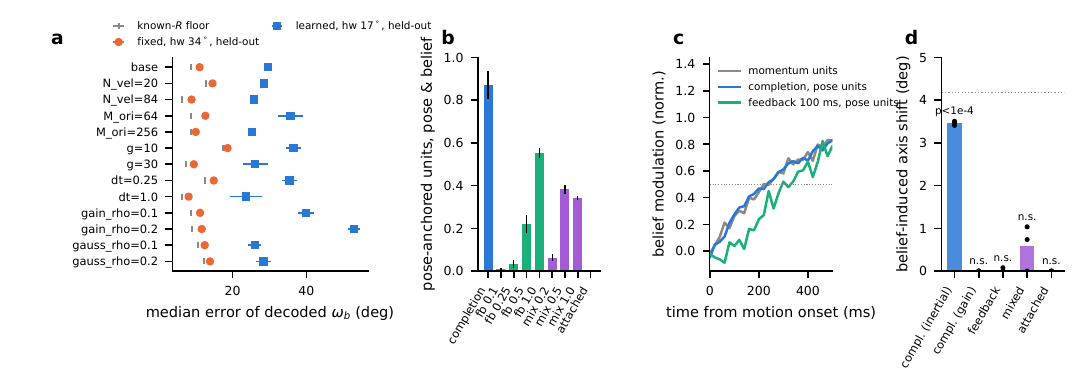}
\caption{Robustness and alternative architectures for the neural model. (a)~Decoding error for each
setting. (b)~Fraction of pose-tuned units modulated by both pose and belief, for the completion, an
attached module, attached modules with feedback of increasing strength and with mixed
selectivity. (c)~Timing of belief modulation in pose-tuned units relative to the physics module's units
for several feedback delays. (d)~Rotation of pose-tuned units' preferred axes by the belief.}
\label{fig:neuralrobust}
\end{figure}

Varying one factor at a time (three seeds each), the fixed readout's error at held-out orientations
stays within $0.5^\circ$ of its error at fitted orientations and $1$--$4^\circ$ above the error of an
observer that knows the orientation, for $20$--$84$ velocity units, $64$--$256$ orientation units,
modulation depths of $10$--$30$~spikes/s, counting windows of $0.25$--$1$~s, shared-gain or
limited-range noise correlations up to $0.2$, and multiplicative noise on the readout weights up to a
relative standard deviation of $0.3$. Co-rotation errors stay at $2$--$5^\circ$. The learned readout
always degrades at held-out orientations with narrow tuning ($24$--$53^\circ$), most under shared-gain
noise. Identification of the object from the momentum population rises with the number of recorded
units ($61\%$ with $10$, $80\%$ with $84$) while the object-agnostic control stays at chance
(Fig.~\ref{fig:neuralrobust}).

\section*{Statements and declarations}
\textbf{Funding.} The author received no funding for this work.
\textbf{Competing interests.} The author is a founder of Kinous Limited, which develops robot sensing and control.
\textbf{Data and code availability.} The code implementing the completion, the estimator and all
experiments, with the configurations and result files needed to reproduce every figure and table, will
be released in a public \texttt{kineframe} repository on publication. Object meshes are the YCB set
\citep{calli}; trajectories are generated with MuJoCo \citep{mujoco}.
\textbf{Ethics approval.} Not applicable: the study involves no human or animal subjects.
\textbf{Author contributions.} Not applicable: S.~Goria is the sole author.

\end{document}